\documentclass[sn-mathphys,Numbered]{sn-jnl}% Math and Physical Sciences Reference Style
\usepackage{graphicx}%
\usepackage{multirow}%
\usepackage{amsmath,amssymb,amsfonts}%
\usepackage{amsthm}%
\usepackage{mathrsfs}%
\usepackage[title]{appendix}%
\usepackage{xcolor}%
\usepackage{textcomp}%
\usepackage{manyfoot}%
\usepackage{booktabs}%
\usepackage{algorithm}%
\usepackage{algorithmicx}%
\usepackage{algpseudocode}%
\usepackage{listings}%
\usepackage{booktabs}

\theoremstyle{thmstyleone}%
\theoremstyle{thmstyletwo}%

\theoremstyle{thmstylethree}%

\begin{document}

\title[Article Title]{BERT4FCA: A Method for Bipartite Link Prediction using Formal Concept Analysis and BERT}


\author[2]{\fnm{Siqi} \sur{Peng}}\email{peng.siqi@iip.ist.i.kyoto-u.ac.jp}
\equalcont{These authors contributed equally to this work.}
\author*[1]{\fnm{Hongyuan} \sur{Yang}}\email{yang.hongyuan.67t@st.kyoto-u.ac.jp}
\equalcont{These authors contributed equally to this work.}

\author[2]{\fnm{Akihiro} \sur{Yamamoto}}\email{yamamoto.akihiro.5m@kyoto-u.ac.jp}

\affil*[1]{\orgdiv{Graduate School of Informatics}, \orgname{Kyoto University}, \orgaddress{\street{Sakyo Ward}, \city{Kyoto}, \postcode{606--8501}, \state{Kyoto}, \country{Japan}}}
\affil[2]{\orgdiv{Graduate School of Informatics}, \orgname{Kyoto University}, \orgaddress{\street{Sakyo Ward}, \city{Kyoto}, \postcode{606--8501}, \state{Kyoto}, \country{Japan}}}


\abstract{We propose BERT4FCA, a novel method for link prediction in bipartite networks, using formal concept analysis (FCA) and BERT. Link prediction in bipartite networks is an important task that can solve various practical problems like friend recommendation in social networks and co-authorship prediction in author-paper networks. Recent research has found that in bipartite networks, maximal bi-cliques provide important information for link prediction, and they can be extracted by FCA. Some FCA-based bipartite link prediction methods have achieved good performance. However, we figured out that their performance could be further improved because these methods did not fully capture the rich information of the extracted maximal bi-cliques. To address this limitation, we propose an approach using BERT, which can learn more information from the maximal bi-cliques extracted by FCA and use them to make link prediction. We conduct experiments on three real-world bipartite networks and demonstrate that our method outperforms previous FCA-based methods, and some classic methods such as matrix-factorization and node2vec.}




\maketitle

\section{Introduction}

\textit{Bipartite link prediction} is the task of predicting the absence or presence of unobserved links in a \textit{bipartite network}~\cite{linkprediction1,linkprediction2,linkpredicition3}. A bipartite network is a structure consisting of two disjoint sets of nodes and a set of edges where every edge only connects two nodes from different sets. Many real-world relational data can be naturally modeled as bipartite networks where the two sets of nodes represent two groups of \textit{entities} and the edges represent their \textit{links} or \textit{relations}~\cite{bipar3,bipar4}. In real-world bipartite networks, some links may be missing or have not been observed yet~\cite{linkpredicition3,linkprediction1}. This leads to the need to predict whether an unobserved link should be a potential new one and gives rise to the task of bipartite link prediction. 

Currently, there are two types of bipartite link prediction. The first type focuses on predicting the relations between nodes from different node sets. For each pair of nodes from different node sets, it predicts whether there should be a link between them if they are not linked with an edge in the original network. For example, in a chemical-disease interaction database, for a chemical and a disease whose interactions are unobserved, it predicts whether they should interact with each other. The second type focuses on predicting the relations between nodes from the same node sets. For two nodes from the same set that are not connected to the same node in the other set, it predicts whether there should be an unobserved node in the other set that is connected to both nodes. For example, in an author-paper network, for two authors that do not have a co-authorship, it predicts if they will have a new co-authorship in the future~\cite{bipar1,FCA2VEC}. Both types of bipartite link prediction have attracted increasing attention for high practical values~\cite{linkpredicition3,matrix,matrix1,drug}. Various methods have been proposed for both types of bipartite link prediction~\cite{linkprediction2,linkpredicition3,linkprediction4}. While some bipartite link prediction methods utilize external information, such as the properties of nodes, most methods make predictions based only on the structural features of the network. Notable methods that use only the structural features include pure rule-based methods like \textit{Common Neighbors}~\cite{linkprediction5} and node embedding methods such as \textit{node2vec}~\cite{node2vec}. 

Recently, research has found that the \textit{bi-cliques} of a bipartite network represent important structural features and can be useful for bipartite link prediction~\cite{weakclique,missbin,linkpredictionbipaetite,FCAnegative}. A \textit{bi-clique} is a complete sub-network of a bipartite network where every node in the first set is linked to every node in the second set. Bi-cliques represent clusters of strongly related entities. For example, in an author-paper network, a bi-clique represents a group of co-researchers and their publications. To capture the information of bi-cliques, research in~\cite{FCA2VEC} proposed \textit{object2vec} and \textit{attribute2vec}, which aims to embed the nodes of a bipartite network into a vector space based on their co-occurrence relationship in the maximal bi-cliques of the network. To achieve this, they used the method of \textit{formal concept analysis} (FCA)~\cite{FCA2VEC,boa,FCAbiclique}. FCA is a method for learning rules from a binary relational knowledge base, which is strongly connected with bipartite networks and bi-cliques. It aims at extracting and organizing \textit{formal concepts} from a \textit{formal context}. A \textit{formal context} represents a collection of binary data of two disjoint sets of entities called \textit{objects} and \textit{attributes} and their binary relation. A \textit{formal concept} represents a maximal group of objects that share the same set of attributes. If we regard the sets of objects and attributes in a formal context as the two node sets in a bipartite network, we will find that the binary relation in the formal context will be equivalent to the set of links in the bipartite network, and the formal concepts extracted from the formal context will be equivalent to the maximal bi-cliques extracted from the bipartite network. With such an equivalence, the co-occurrences of nodes in a maximal bi-clique can be converted to the co-occurrences of objects in a formal concept, which resembles the co-occurrences of words in a sentence. Thus, one can use the embedding models similar to the well-known word embedding model \textit{Word2Vec}~\cite{word2vec1,word2vec2} to embed the nodes into vectors~\cite{FCA2VEC}. Besides such a Word2Vec-like embedding method, in~\cite{boa}, the authors proposed another method for embedding the nodes based on bi-clique information with the help of \textit{Bidirectional Long Short-Term Memory} (Bi-LSTM)~\cite{bilstm}. Both methods have shown good performances~\cite{boa,FCA2VEC}. These two methods open up a novel strategy for bipartite link prediction, that is, to convert bipartite networks into formal contexts and use FCA to extract information on bi-cliques so that we can further process them with other methods. However, their methods have two limitations. First, they did not fully utilize all the information that FCA can extract. Second, their methods can only conduct the first type of link prediction, that is, predicting the relations between nodes from the same node set. We consider it better to design a general method that can conduct both types of bipartite link prediction.

To address these limitations, we propose a novel method called \textit{BERT4FCA}, which is designed to capture more information provided by FCA and use such information to make both types of bipartite link prediction with the help of a popular method in natural language processing method called \textit{BERT} or \textit{Bidirectional Encoder Representations from Transformers}~\cite{Transformer}. BERT is a language model that is first pre-trained on unlabeled free text to learn the co-relations between words and sentences and then fine-tuned on a small labeled dataset to fit a target downstream task. We chose BERT because we found that the information provided by FCA, such as the formal concepts and their hierarchical relations, shares some similarities with the input data that BERT takes. Hence, after processing the information provided by FCA and modifying the basic structure of BERT, we will be able to capture all such information during pre-training. Then, by fine-tuning the pre-trained model to fit our target bipartite link prediction tasks, we are expected to get better results than those embedding methods thanks to the extra information we have captured. Additionally, in the field of natural language processing, while Word2Vec requires task-specific networks for downstream tasks, BERT can be fine-tuned for multiple tasks using the same pre-trained model, which is more convenient. We believe the convenience is preserved by our method since we do not need to design two different networks for two link prediction tasks.

Our main contributions are as summarized as follows:
\begin{itemize}
    \item We propose a novel FCA-based method called BERT4FCA for bipartite link prediction, which can capture and learn more information extracted by FCA than previous FCA-based methods. We conduct experiments to show that learning more information given by FCA contributes to higher link prediction performance.
    \item Experimental results show that our method outperforms all previous FCA-based methods as well as classic non-FCA-based methods on both types of bipartite link prediction.
    \item We demonstrate that the information on the order relation between maximal bi-cliques is beneficial for bipartite link prediction. To the best of our knowledge, no research has discovered it before.
    \item To the best of our knowledge, we are the first to provide a general method for using BERT to learn information from the concept lattices provided by FCA. Although in this paper, we only discuss its application in bipartite link prediction, we plan to conduct further research on applying it to other FCA-related tasks.
\end{itemize}

The rest part of this paper is organized as follows. In Section 2, we start with some preliminaries, including bipartite networks, bi-cliques, FCA, and BERT. In Section 3, we present the problem formulation of bipartite link prediction and provide an overview of related work. In Section 4, we introduce and analyze our method BERT4FCA. In Section 5, we describe our experiments on three real-world datasets and discuss the results. Finally, in Section 6, we draw a conclusion and discuss our plans for future work.

\section{Preliminaries}

\subsection{Bipartite Networks and Bi-cliques}

\begin{figure}[!htbp]
    \centering
    \includegraphics[width=\textwidth]{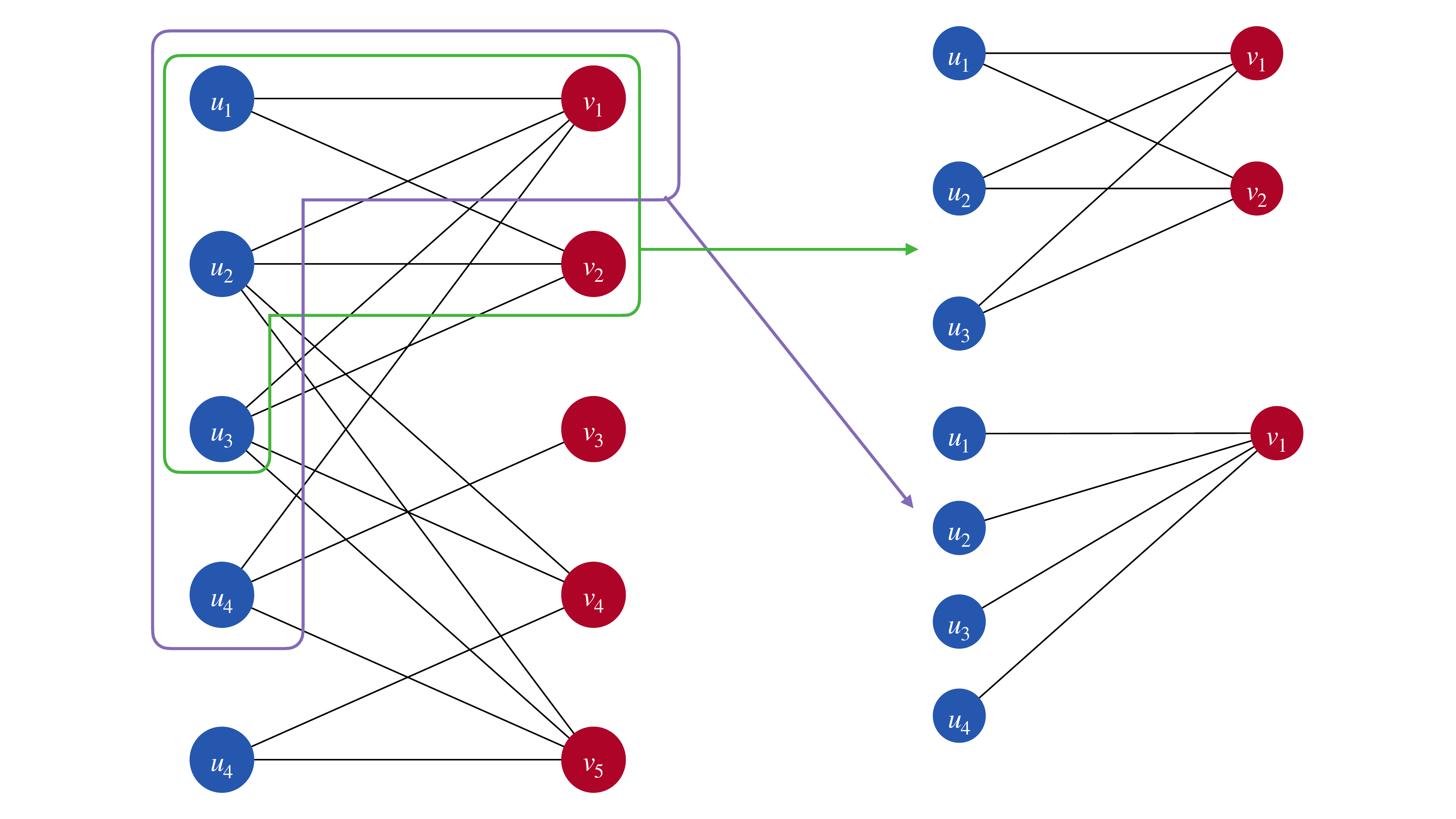}
    \caption{An example of a bipartite network $(U,V,E)$ and two of its bi-cliques. The nodes in blue form the node set $U$, and the nodes in red form the node set $V$. The two sub-networks framed in green and purple are two maximal bi-cliques of the network. }
    \label{bipartitenetwork}
\end{figure}

\textbf{Bipartite Networks}: A \textit{bipartite network} $C$ is a triple $(U, V, E)$ where $U$, $V$ are two disjoint sets of \textit{nodes} and $E$ is a set of \textit{edges}. Each edge connects a node $u\in U$ to another node $v\in V$ and is denoted as $(u,v)$. That is, we should have $U\cap V=\emptyset$ and $E\subseteq U\times V$. Fig.~\ref{bipartitenetwork} gives an example of a bipartite network, where all nodes $u$ in $U$ are colored blue, and all nodes $v$ in $V$ are colored red. We can observe that edges only exist between pairs of nodes from different sets of nodes. 

\textbf{Sub-networks}: A bipartite network $C_1 = (U_1, V_1, E_1)$ is a \textit{sub-network} of $C=(U,V,E)$ if both sets of nodes and the set of edges of $C_1$ are subsets of those of $C$, \textit{i.e.}, $U_1 \subseteq U, V_1 \subseteq V$, and $E_1 \subseteq E$.

\textbf{Bi-cliques}: A bipartite network $C_1 = (U_1,V_1,E)$ is a \textit{bi-clique} of $C = (U,V,E)$ if $C_1$ is a \textit{sub-network} of $C$ and there is an edge between every node pair from different subsets of nodes in $C_1$, \textit{i.e.}, $U_1 \subseteq U, V_1 \subseteq V, E_1 \subseteq E$ and $U_1\times V_1=E_1$.

\textbf{Maximal Bi-cliques}: A bi-clique $C_1 = (U_1,V_1,E_1)$ is a \textit{maximal bi-clique} of a bipartite network $C = (U,V,E)$ if it is not a sub-network of any other bi-cliques of the bipartite network, \textit{i.e.}, $\forall C_2 = (U_2,V_2,E_2)$ such that $U_1 \subseteq U_2 \subseteq U$ and $V_1 \subseteq V_2 \subseteq V$, $U_2\times V_2=E_2$ is satisfied if and only if $U_1 = U_2, V_1 = V_2$, and $E_1 = E_2$. Figure 1 gives some examples of maximal bi-cliques. 

\subsection{Formal Concept Analysis (FCA)}
\textit{Formal concept analysis} (FCA) is a method for learning rules from a binary relational knowledge base called a \textit{formal context}. Given a formal context, it aims to extract \textit{formal concepts} and their hierarchical structures, which constitute \textit{concept lattices}. In this paper, we only briefly introduce the minimal necessary notions of FCA. For a detailed introduction, please refer to~\cite{FCAbook,FCAintroduction,FCAsurvey1,FCAsurvey2}. 

\textbf{Formal Contexts}: A formal context is a triple $\mathbb K :=(G,M,I)$, where $G$ is a set of \textit{objects}, $M$ is a set of \textit{attributes}, and $I\subseteq G \times M$ is a binary relation called \textit{incidence} that expresses which object has which attribute. We write $gIm$ or $(g,m)\in I$ to express that the object $g\in G$ has the attribute $m\in M$.

\begin{figure}[!htbp]
    \centering
    \includegraphics[width=\textwidth]{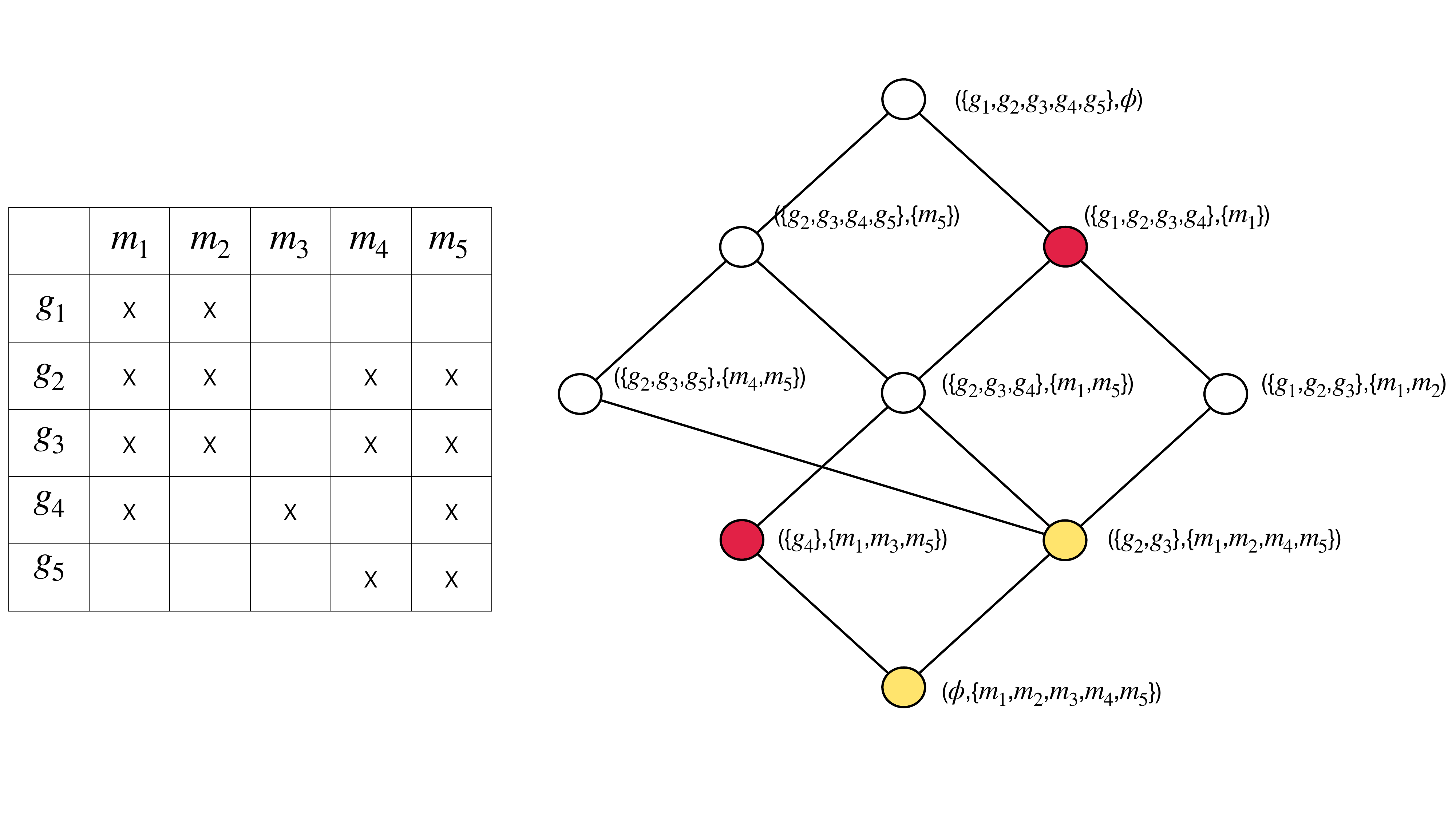}
    \caption{Left: A sample formal context. Right: The concept lattice corresponding to the formal context in the left panel. The nodes in yellow are neighbors and the nodes in red are not neighbors. }
    \label{conceptlattice}
\end{figure}

Formal contexts are illustrated in binary tables, as exemplified in the left of Fig.~\ref{conceptlattice}, where rows correspond to objects and columns to attributes, and a cell is marked with a cross if the object in its row has the attribute in its column. In the context shown in the left panel of Fig.~\ref{conceptlattice}, the marked cell represents that the object listed in the row possesses the corresponding attribute in the column.

\textbf{Formal Concepts}: In a context $\mathbb K =(G,M,I)$, for subsets of objects and attributes $A \subseteq G$ and $B \subseteq M$, $(A,B)$ is called a \textit{formal concept} if $\forall (A_1, B_1)$ such that $A\subseteq A_1\subseteq G$ and $B\subseteq B_1\subseteq M$, $A_1\times B_1 \subseteq I$ is satisfied if and only if $A=A_1$ and $B=B_1$. If $(A, B)$ is a formal concept, $A$ is also called an \textit{extent}, and $B$ is also called an \textit{intent}.

\textbf{Concept Lattices}: Given a context $\mathbb{K} = (G, M, I)$, the \textit{concept lattice} of context $\mathbb{K}$, denoted by $\underline {\mathfrak B}(\mathbb{K})$, is the structure that organizes the set of all concepts extracted from context $\mathbb{K}$ with the \textit{hierarchical order} $<$. For two concepts $(A_1,B_1) $ and $(A_2,B_2)$, we write $(A_1,B_1) < (A_2,B_2)$ if $A_1\subset A_2$ (which mutually implies $B_2\subset B_1$).

Concept lattices are usually figuralized with line diagrams. For example, the line diagram shown in the right panel of Fig.~\ref{conceptlattice} represents the concept lattice of the context represented in the left panel of the same figure. In the diagram, nodes represent formal concepts and lines represent hierarchical orders.

\textbf{Neighbor of Concepts}: In a concept lattice $\underline{\mathfrak B} (\mathbb K)$, for two concepts $(A_1, B_1)$ and $(A_2, B_2)$ such that $(A_1, B_1) < (A_2, B_2)$, $(A_1, B_1)$ is defined to be the \textit{lower neighbor} of $(A_2, B_2)$ if $\nexists (A_3, B_3)\in \underline{\mathfrak B} (\mathbb K)$ such that $(A_1, B_1) < (A_3, B_3) < (A_2, B_2)$~\cite{FCAbook}. In this case, $(A_2, B_2)$ is dually called the \textit{upper neighbor} of $(A_1, B_1)$ or we may also directly say that $(A_1, B_1)$ and $(A_2, B_2)$ are \textit{neighbors} or that they have the \textit{neighboring relation}.

In the line diagram representation of a concept lattice, if two concepts are neighbors, they will be shown as two nodes directly connected by a line. For example, in the concept lattice shown on the right of Fig.~\ref{conceptlattice}, the nodes in yellow are neighbors, and the nodes in red are not neighbors.

\subsubsection{FCA and Bipartite Networks}

\begin{figure}
    \centering
    \includegraphics[width=\textwidth]{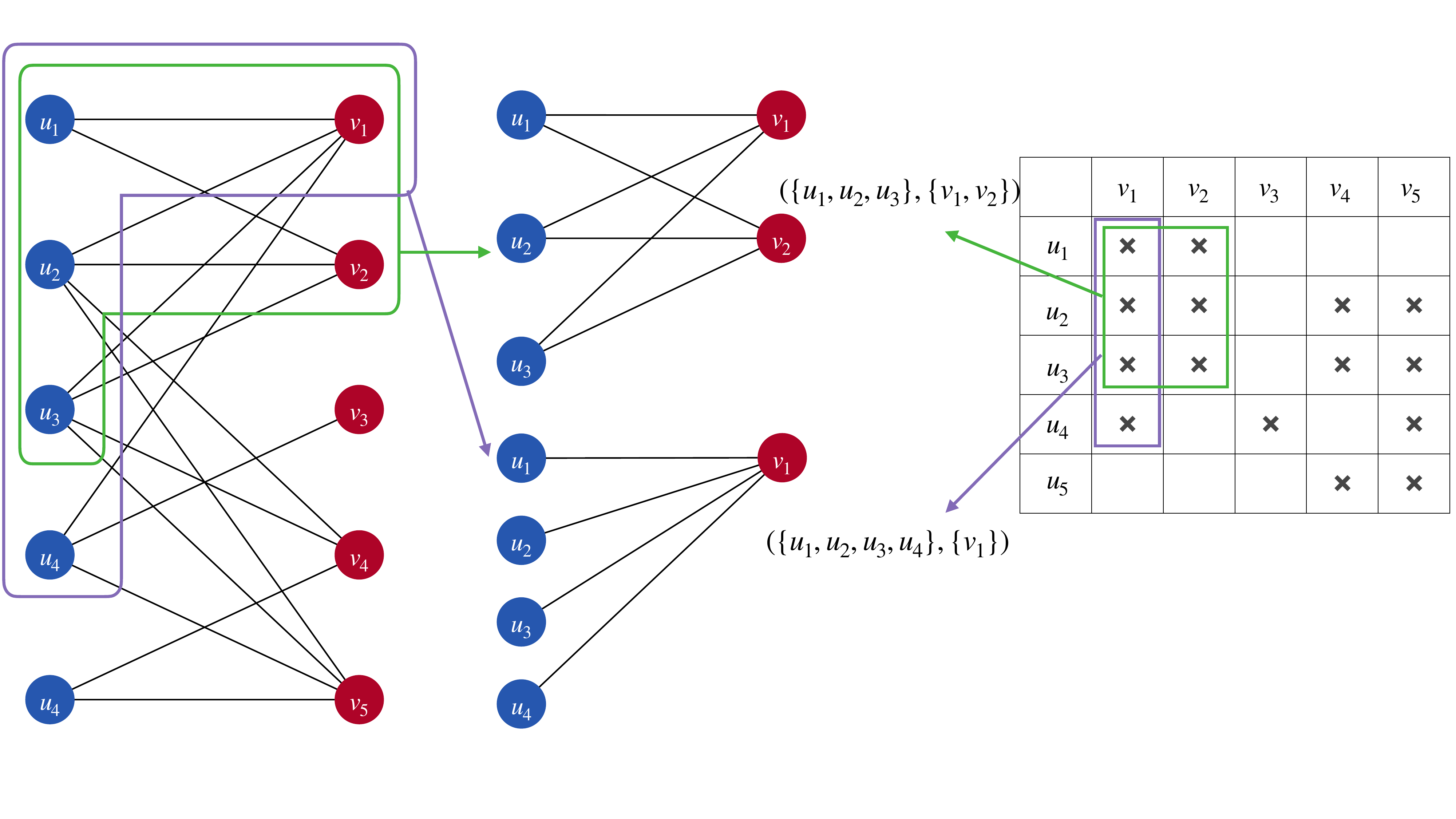}
    \caption{A depiction of the equivalence between bipartite networks and formal contexts, as well as the equivalence between maximal bi-cliques and formal concepts. The bipartite network to the left can be represented as the formal context to the right. The sub-network circled in purple and green are maximal bi-cliques in the bipartite networks to the left, which can be represented into two formal concepts framed in the corresponding colors in the formal context to the right.}
    \label{FCAbiclique}
\end{figure}

By comparing the definitions above, we can easily find that the definition of a formal context is equivalent to that of a bipartite network, and the definition of a maximal bi-clique is equivalent to that of a formal concept. That is, for every bipartite network $C = (U, V, E)$, if we consider the two node sets of a bipartite network $U$ and $V$ as the object set and the attribute set, and the edge set $E$ as the binary relation of the objects and attributes, we may easily find that $(U, V, E)$ should also be a formal context. Also, if $C_1=(A_1, B_1, E_1)$ is found to be a maximal bi-clique of $C$, it is certain that $(A_1, B_1)$ is also a formal concept in $\underline{\mathfrak B} (U, V, E)$. Fig.~\ref{FCAbiclique} gives an example of such an equivalence between bipartite networks and formal contexts, as well as the equivalence between maximal bi-cliques and formal concepts.

Many previous research has spotted and discussed the two types of equivalences. However, besides these two types of equivalences, we may also find that the order relation and neighboring relation between formal concepts also correspond to the cover relations between maximal bi-cliques. The hierarchical order and neighboring relation between formal concepts are considered important in FCA and are represented by concept lattices, so in bipartite networks, the cover relations between maximal bi-cliques may also offer valuable insights for understanding the networks' structures. However, to the furthest of our knowledge, no previous studies on bipartite network theory have discussed on it. Hence, we believe that if we can design a model for learning the cover relations between bi-cliques, it will hopefully capture more information on the networks' structure and thus have a better performance in bipartite link prediction.
\subsection{BERT}

\textit{BERT}, short for \textit{Bidirectional Encoder Representations from Transformers}~\cite{BERT}, is a language model based on the \textit{Transformer} architecture~\cite{Transformer}. It works in a ``\textit{pre-train} first and \textit{fine-tune} next'' mode. First, it pre-trains a large language model with two general tasks on a large amount of unlabeled free text. Then, it fine-tunes the pre-trained model with a specific downstream task on a small labeled dataset. 

The two general tasks used in the pre-training phase are \textit{mask language model} (MLM) and \textit{next sentence prediction} (NSP), which are defined as follows.

\textbf{MLM}: MLM is the task that predicts the full sentence from a sentence where some words are randomly masked with a special token ``[MASK]''. For example, if we have a sentence like ``the quick brown fox jumps over the lazy dog'', the model will take a masked version of the sentence like ``the [MASK] brown fox jumps [MASK] the lazy dog'' as input and the original sentence as target for output. The task helps the model learn the co-occurrence relationship between words in the same sentence~\cite{BERT}.

\textbf{NSP}: NSP is the task that predicts whether two sentences are the subsequent sentences or not. The model takes a sentence pair $(A, B)$ as input and is expected to output TRUE if $B$ is the sentence directly following $A$ in the corpus, or FALSE otherwise. The task helps the model to understand the relationships between sentences. 

After pre-training, the values of the weight matrices of the pre-trained model are used as initial values for that of the fine-tune model and updated in the fine-tuning phase~\cite{BERT2}. This is possible because pre-training and fine-tuning should use exactly the same network architecture except for the output layer. The downstream task in the fine-tuning phase should be the final target task. Hence, after fine-tuning, we get the final model which can be directly used for our target task.

\subsubsection{BERT and FCA}
As mentioned above, BERT is a language model, which takes sentences in natural languages as inputs. Although formal concepts are not sentences in natural languages, they do share some similarities -- If we regard objects and attributes as words, then the extents and intents of a concept can be regarded as sentences in a language that has an order-free syntax. With such similarities, we may expect that the pre-training phase of BERT, which was originally designed for learning the features of words and sentences, can also be used for learning the features of objects, attributes, and formal concepts from a formal context and its corresponding concept lattice. We may also expect that bipartite link prediction can be a suitable downstream task after such a pre-training because the information of nodes (\textit{a.k.a.} objects and attributes), and maximal bi-cliques (\textit{a.k.a.} formal concepts) learned in the pre-training phase is helpful in increasing the accuracy of predictions.

Note that in most natural languages, the syntax is not order-free, indicating that different word orderings usually have completely different meanings. Hence, BERT has a special mechanism called \textit{position embedding} for capturing the order of words in the input sentence. However, since the extents and intents of formal concepts are unordered, there is no need to keep track of the order. Hence, in our method, this mechanism is removed from BERT.

\section{Problem Formulation and Related Work}

\subsection{Problem Formulation}
The research deals with the problem of bipartite link prediction, which consists of two different tasks -- the first task predicts the missing or unknown relation between two nodes from the same node set of a bipartite network, and the other task predicts the missing or unknown relation between two nodes from different node sets. As analyzed above, a bipartite network is equivalent to a formal context, and the two node sets are equivalent to the object set and the attribute set. Hence, for ease of understanding, we hereby name the two tasks as the \textit{object-object task} or the \textit{O-O task} and the \textit{object-attribute task} or the \textit{O-A task}. The two node sets of a bipartite network are also directly called the \textit{objects} and \textit{attributes} in the rest part of the paper. 

The formal definitions of the two tasks are defined as follows.

\textbf{O-O task}: Given an original network $C=(U,V,E)$ and a target network $C'=(U,V',E')$, the O-O task aims to predict if a group of nodes $G=\{u_1,u_2,\cdots,u_{|G|}\}\subseteq U$ that does not have \textit{object-object links} in $C$, should have object-object links in $C'$. For a group of nodes $G\subseteq U$, they are considered to have \textit{object-object links} in a network $(U,V,E)$ if $\exists v\in V$ such that $\forall u\in G$, $(u,v)\in E$.

\textbf{O-A task}: Given an original network $C=(U,V,E)$ and a target network $C'=(U,V,E')$, the O-A task aims to predict if two nodes $u\in U$ and $v\in V$ that does not have an \textit{object-attribute link} in $C$, should have an object-attribute link in $C'$. For two nodes $u\in U$ and $v\in V$, they are considered to have an \textit{object-attribute link} in a network $(U,V,E)$ if $(u,v)\in E$.

\subsection{Related Work on FCA-based Bipartite Link Prediction}

Previous methods for FCA-based bipartite link prediction can be classified into two groups -- rule-based methods and embedding-based methods. The rule-based methods directly use pre-determined rules to make predictions based on the information provided by FCA; the embedding-based methods embed the objects and attributes into vectors using the information provided by FCA and use the embedding vectors for making link predictions.

In~\cite{missbin}, the authors proposed a rule-based method for the O-A task by analyzing the overlapping formal concepts from a formal context. If the ratio of the overlapped part exceeds a predefined threshold, they are considered parts of the same large formal concept, so all the missing links in the non-overlapped part \textit{a.k.a.} \textit{the structure hole}~\cite{structualhole} are predicted as present links. It presents an interesting approach to bipartite link prediction but only has a limited performance. In~\cite{matrixnega}, the authors also used this approach, but only as a pre-processing step before the main link prediction step using matrix factorization. This hybrid method has much better results than the pure rule-based method.

In~\cite{FCA2VEC}, the authors proposed an embedding-based method called \textit{object2vec} to embed objects into vectors~\footnote{Dually, they have also proposed \textit{attribute2vec} for embedding attributes into vectors, which uses exactly the same mechanism as object2vec.} using the information from the formal concepts. It has two embedding models, \textit{object2vec-CBoW} and \textit{object2vec-SG}, both are derived from \textit{Word2Vec}~\cite{word2vec1,word2vec2}. Object2vec-CBoW, based on the \textit{continuous-bag-of-words} model from Word2Vec, predicts a target object using objects around it within the same extent; object2vec-SG, based on the \textit{skip-gram} model from Word2Vec, uses an object to predict other objects in the same extent. They conducted experiments on the O-O task on an author-publication network and demonstrated good performance. 

In~\cite{boa}, the authors proposed another embedding method called \textit{Bag of Attributes (BoA)}. Their method trains a more complicated embedding model using \textit{Bidirectional Long Short-Term Memory} (Bi-LSTM)~\cite{bilstm} and \textit{Variational Autoencoder} (VAE)~\cite{vae} on formal contexts. They conducted experiments on the O-O task on the same datasets as object2vec, and the results are similar to those obtained with object2Vec. 

Both rule-based methods and embedding-based methods have some limitations. The rule-based methods use pre-determined rules, which may only work on some specific types of networks; the embedding-based methods use the information provided by FCA to train embedding vectors, but their model did not make full use of all the information provided by FCA -- for example, the ordering relation and neighboring relation of formal concepts are not learned into the embeddings. 

\section{The Proposed Method}\label{sec4}
To address the limitations of the previous FCA-based methods for bipartite link prediction, we propose a novel method named \textit{BERT4FCA}, which is designed to better learn the information of concept lattices and can conduct both the O-O task and the O-A task. We name our method \textit{BERT4FCA} because it provides a general framework for using BERT to learn and utilize the information of concept lattices, so it is expected to be generally applicable to all tasks related to FCA. Although in this research, we only discuss its application in the two bipartite link prediction tasks, we plan to study the possibility of applying our method to other tasks in the future.

%In this section, we will introduce our method, BERT4FCA, and analyze why it can outperform previous methods. Notice that in FCA, objects in extents and attributes in intents are dual. Therefore, in the rest of this section, we will refer only to objects and extents. Attributes and intents will be treated the same. 

Similar to all previous FCA-based bipartite link prediction methods, the objective of our method is to learn information from concept lattices and use it to make two link prediction tasks. However, the information our method aims to learn not only includes the extent and intent of a formal concept, but also includes the neighboring relations between formal concepts. Note that here we choose to learn the \textit{neighboring relations} instead of the general \textit{order relations} because, to reconstruct the concept lattice, the neighboring relations are enough~\cite{FCAbook}.

\begin{figure}
    \centering
    \includegraphics[width=\textwidth]{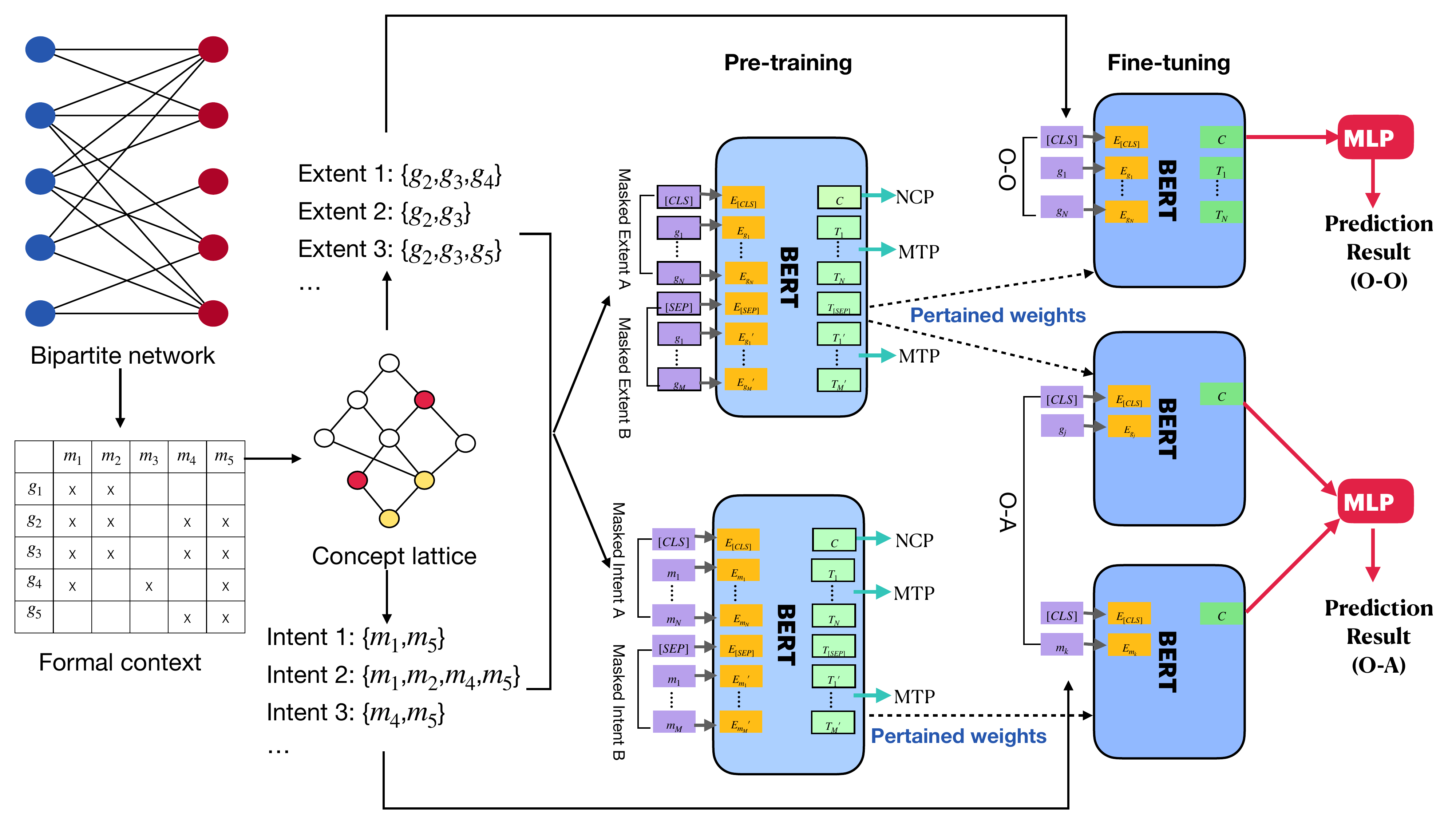}
    \caption{An overview of the working flow of our method.}
    \label{BERT4FCA}
\end{figure}

Our method consists of 4 steps: \textit{data preparation with FCA}, \textit{input tokenization}, \textit{BERT pre-training}, and \textit{BERT fine-tuning}. An overview of the workflow of our method is shown in Fig.~\ref{BERT4FCA}. 

\textbf{Data Preparation}: In this step, we convert the bipartite network into a formal context, extract all formal concepts, and construct the concept lattice. Then, we extract the neighboring relations between concepts from the concept lattice. We use Z-TCA\footnote{The algorithm is originally designed for \textit{triadic concept analysis} (TCA), the 3-dimensional generalization of FCA, but can also be used for FCA.}\cite{ztca} for extracting all formal concepts from a formal context, and use a \textit{topological sorting} algorithm to extract neighboring relations from the concept lattice. The details of the algorithm are presented in Algorithm~\ref{tpsort}. After this step, we will obtain all extents and intents and neighboring concepts in the concept lattice. 

\begin{algorithm}[ht]
\caption{Get all neighboring relations using topological sorting.}
\hspace*{\algorithmicindent} \textbf{Input} A formal context $\mathbb{K}=(U,V,E)$ and its concept lattice $\underline{\mathfrak B}\mathbb{(K)}$.\\
\hspace*{\algorithmicindent} \textbf{Output} $\{\mathrm{N}(C)\}$, a list of the lower neighbors of concepts. Here $\mathrm{N}(C)$ represents the lower neighbor of $C\in (\underline{\mathfrak B}\mathbb{(K)} - (\emptyset, \emptyset''))$.
\begin{algorithmic}[1]
\For{$C\in \underline{\mathfrak B}\mathbb{(K)}$}
    \State{$\mathrm{D}(C) \leftarrow |\{C_1\in \underline{\mathfrak B}\mathbb{(K)}~|~C_1<C\}|$}
\EndFor
\State{Create $Q$ as an empty queue.}
\For{$C\in \underline{\mathfrak B}\mathbb{(K)}$}
    \If{$\mathrm{D}(C)=0$} 
        \State{Push $C$ into the back of $Q$.}
    \EndIf
\EndFor
\While{$Q$ is not empty}
    \State{Fetch $C$ from the front of $Q$ and pop it.}
    \For{$C_1\in \underline{\mathfrak B}\mathbb{(K)}$ such that $C < C_1$}
        \State{$\mathrm{D}(C_1)\leftarrow \mathrm{D}(C_1) - 1$}
        \If{$\mathrm{D}(C_1) = 0$}
            \State{$\mathrm{N}(C_1)\leftarrow C$}
            \State{Push $C_1$ into the back of $Q$.}
        \EndIf
    \EndFor
\EndWhile

\end{algorithmic}
\label{tpsort}
\end{algorithm}

\textbf{Input Tokenization}: In this step, the objects and attributes are tokenized into one-hot vectors. These one-hot vectors are further converted into dense vectors through the \textit{input embedding} so that they can be processed by BERT. 

The input embedding is the sum of two parts: \textit{tokenization embedding} and \textit{segment embedding}. Tokenization embedding is the general-sense embedding that uses a full-connect layer to embed the tokenized one-hot vectors into a dense vector space. Segment embedding is used to embed the tokens in the segment-info sequence into the vector space of the same dimension as the tokenization embedding. In BERT, sometimes the training sample may be a concatenation of two different sequences (details will be introduced later). In this case, the segment-info sequence is generated to distinguish the two different sequences within the same training sample. For example, in a training sample, the input is a sequence of seven tokens, where the first four tokens belong to the first sequence, and the last three tokens belong to the second sequence. Then, the segment-info sequence of this training sample should be $(0,0,0,0,1,1,1)$.

Note that, as mentioned above, in the original framework of BERT, there is another input embedding called \textit{position embedding} used for learning the order of words in a sentence. In BERT4FCA, however, it is removed because the input sequences are all unordered.

\textbf{BERT Pre-training}: In this step, we use the BERT framework to pre-train two models -- the \textit{object model} and the \textit{attribute model} -- on all extents and intents, correspondingly. Here, we only introduce the pre-training process of the object model for example. The input of a training sample of the pre-training of the object model is a pair of sequences, each representing an extent. Since the lengths of extents may be different, we pad short extents to make all extents have the same length with a special token ``[PAD]''. We also add a special token, ``[SEP]'', between two sequences in order to separate them.

The model is trained with two tasks: \textit{masked token prediction} (MTP) and \textit{neighboring concepts prediction} (NCP), derived from the MLM and NSP tasks in the original version of BERT, correspondingly. 

MTP is the task that helps the model learn the co-occurrence relationships between objects within the same extent. In the task, we randomly select a certain percentage of objects in both extents to be masked. For each object to be masked, we replace them into a special token ``[MASK]'' with 80\% probability, or replace it with a random object with 10\% probability, or keep it unchanged with 10\% probability. Then, the model takes the masked pair of extents as input and the unmasked pair of extents as the target for output. That is, it is trained to predict the masked objects in the extents. 

NCP is the task that helps the model learn the neighboring relations between formal concepts in the concept lattice. In the task, the model takes a pair of extents as input and is expected to output TRUE if the pair of extents corresponds to a pair of formal concepts that have neighboring relations or output FALSE otherwise. Note that in this task, clearly, the number of negative samples is much larger than the positive samples. To get a balanced training set, we randomly select a small portion of negative samples and keep the number of positive and negative samples to be the same.

The two tasks are trained simultaneously -- the training loss is the sum of the losses of the two tasks. After pre-training, the pre-trained models are expected to have captured and stored information on the concept lattice, including relationships between objects and formal concepts.

\textbf{BERT Fine-tuning}: In this step, we fine-tune the pre-trained object model and attribute model to make them fit our target bipartite link prediction tasks. 

For the O-O task, we fine-tune the pre-trained object model with the training samples generated from the original bipartite network $(U,V,E)$. The training samples are generated with the following steps. First, we determine $l_m$, the maximal length of a group of objects we want to predict. Then, we enumerate all object subsets $U_1\subseteq U$ such that $|U_1|\leq l_m$. For each $U_1$, we create a training sample that takes the tokenized and padded sequence of $U_1$ as input, and the label for the sample is set to TRUE if the objects in $U_1$ have object-object links and FALSE otherwise. After generation, the training samples are fed into a network that has the same structure as the network used in the pre-training step, except that in the last layer, the hidden states are not fed into the output layer used for the two pre-training tasks, but are fed into an output layer specified to the O-O task. Suppose the final output of the fine-tuning network for the O-O task is $P_\mathrm{O-O}$; the last hidden state of the basic BERT network, \textit{i.e.}, the ``[CLS]'' representation~\footnote{For more details on the network structure of BERT, please refer to~\cite{BERT}.} is $\boldsymbol{h}_\mathrm{L}^\mathrm{[CLS]}$, then the O-O-task-specified output layer can be described as follows:
\begin{equation}
P_{\mathrm{O-O}} = \sigma(\mathrm{ReLU}(\boldsymbol{h}_\mathrm{L}^\mathrm{[CLS]}\boldsymbol{W}_\mathrm{CLS})\boldsymbol{W}),
\end{equation}
where $\sigma(\cdot)$ is the sigmoid function; $\mathrm{ReLU}(\cdot)$ is the \textit{Rectified Linear Unit} (ReLU)~\cite{relu}; $\boldsymbol{W}_{\mathrm{[CLS]}}$ and $\boldsymbol{W}$ are weight matrices.

For the O-A task, we fine-tune both the pre-trained object model and the attribute model together with training samples generated from the original bipartite network. The training samples are generated with the following steps. We first enumerate all objects $u\in U$ and $v\in V$. For each pair of $(u,v)$, we create a training sample that takes the tokens of $u$ and $v$ as input, and the label for the sample is set to TRUE if $(u,v)\in E$ and FALSE otherwise. After generation, the tokenized object and tokenized attribute are fed into two separate BERT networks -- except for the last output layer, the first network has the same structure as the network used in pre-training the object model, and the second network has the same structure as that used in pre-training the attribute model. The last hidden states of both networks are concatenated and fed through a single output layer specified for the O-A task. Suppose the final output of the fine-tuning network for the O-A task is $P_\mathrm{O-A}$; the last hidden states of the two BERT networks are $\boldsymbol{h}_\mathrm{L1}^\mathrm{[CLS]}$ and $\boldsymbol{h}_\mathrm{L2}^\mathrm{[CLS]}$, then the O-A-task-specified output layer can be described as follows:
\begin{equation}
    P_{\mathrm{O-A}} = \sigma(\mathrm{ReLU}({(\boldsymbol{h}_\mathrm{L1}^\mathrm{[CLS]}}^\frown \boldsymbol{h}_\mathrm{L2}^\mathrm{[CLS]})\boldsymbol{W}_\mathrm{CLS})\boldsymbol{W}),
\end{equation}
where $\boldsymbol{a}^\frown \boldsymbol{b}$ means the concatenation of vectors $\boldsymbol{a}$ and $\boldsymbol{b}$.

\begin{figure}[!htbp]
    \centering
    \includegraphics[width = \textwidth]{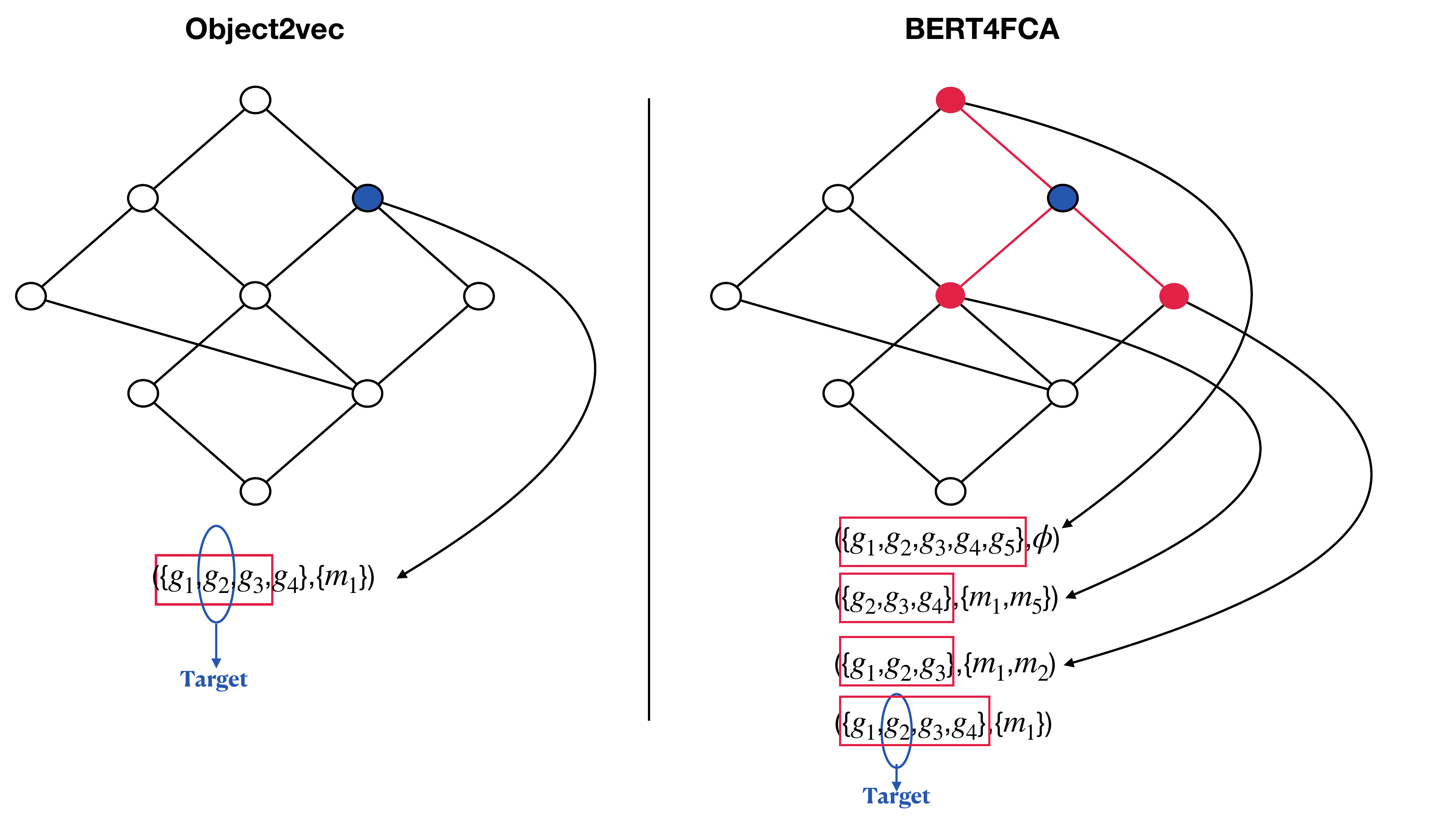}
    \caption{ The comparison of how much information from a concept lattice is learned and used when predicting an object by two methods, with object2vec shown on the left and BERT4FCA shown on the right. The target object to be predicted is circled in blue. The information used for predicting the object is shown in red.}
    \label{FCA2VEC}
\end{figure}

Above is the whole workflow of our method. From above, we can see that our method has two advantages. First, in the pre-training step, our model can learn more information from concept lattices compared to the previous FCA-based embedding methods like object2vec. As shown in Fig.~\ref{FCA2VEC}, in object2vec, when embedding an object, it uses only the information of a small set of objects within the same extent. In the pre-training step of our method, however, we learn the feature of an object using the information from all objects in the entire extent as well as the objects in the extents of its neighbor concepts in concept lattice. Second, our method works in a pre-train first and fine-tune next mode that first pre-trains two large models and then fine-tunes them on various downstream tasks related to formal context, meaning that if we wish to conduct new tasks on a formal context where we have pre-trained the object model and attribute model, we can skip the first three steps and directly conduct the final fine-tuning step. Previous methods, however, may need to re-train the models when coming to a new task.

\section{Experiments}\label{sec5}

\subsection{Datasets}\label{subsec5}
We conduct experiments on three real-world datasets: \textit{ICFCA}, \textit{BMS-POS}, and \textit{Keyword-Paper}. All three datasets are used for both the O-O task and the O-A task. We depict the features of these datasets in Table~\ref{tab:feature}. A detailed description of each follows. 

\begin{table}[!htbp]
    \centering
    \begin{tabular}{ccccccc}
    \hline
         Dataset                &  Task                  & Input/Target &  Objects&  Attributes&  Edges&  Concepts\\
    \hline
         \multirow{4}{*}{ICFCA}         &  \multirow{2}{*}{O-O}  & Input   & 334 & 12614 & 13399 & 775 \\
                                        &                        & Target       & 334 & 12614 & 15980 & 844 \\
                                        &  \multirow{2}{*}{O-A}  & Input        & 351 & 12614 & 14445 & 878 \\
                                        &                        & Target       & 351 & 12614 & 16049 & 922 \\
         \multirow{4}{*}{BMS-POS}       &  \multirow{2}{*}{O-O}  & Input        & 468 & 1946 & 7376 & 7791  \\
                                        &                        & Target       & 468 & 1946 & 8085 & 10235 \\
                                        &  \multirow{2}{*}{O-A}  & Input        & 468 & 1946 & 7376 & 7791  \\
                                        &                        & Target       & 468 & 1946 & 8085 & 10235 \\
         \multirow{4}{*}{Keyword-Paper} &  \multirow{2}{*}{O-O}  & Input        & 162 & 5640 & 7274 & 1610 \\
                                        &                        & Target       & 162 & 5640 & 8308 & 2049 \\
                                        &  \multirow{2}{*}{O-A}  & Input        & 162 & 5206 & 7648 & 1713 \\
                                        &                        & Target       & 162 & 5206 & 7907 & 2046 \\
    \hline
    \end{tabular}
    \caption{The features of the three datasets.}
    \label{tab:feature}
\end{table}
\textbf{ICFCA}: The ICFCA dataset is an author-paper network provided by~\cite{FCA2VEC} -- the objects represent the authors, the attributes represent the publications, and each edge (\textit{a.k.a.} relation) represents the author is in the author list of the publication. This dataset is generated from \textit{Digital Bibliography \& Library Project} (DBLP) dump on 1st Aug 2019 which is available at https://dblp.uni-trier.de/xml/. 

For the O-O task, we are to simulate the practical case in which we wish to predict future co-authorships or seek potential co-authors from an author-paper network at a certain time point. Hence, we generate a history network from the full network as the input network for this task by removing the authors, publications, and author-paper edges after 1st Jan 2016; we generate the current network as the target network for this task from the full network by removing authors who had no publication before 31st Dec 2015 and their corresponding edges.

For the O-A task, we are to simulate the practical case that some parts of the network are missing, and we wish to use the known edges in the network to predict the potentially missing edges. We generate the input network for this task from the full network by randomly removing 10\% of author-paper edges; we use the full network as the target network for this task.

\textbf{BMS-POS}: The BMS-POS dataset is a product purchased transactions network provided by KDD Cup 2000 -- the objects represent the products, the attributes represent purchasing transactions, and an edge represents that the product is bought in a certain purchasing transaction. The original data is very large, so in this research, we only select the first 1946 transactions. The dataset is available at https://kdd.org/kdd-cup/view/kdd-cup-2000. 

For the O-O task, we are to simulate the practical case in which we wish to predict two products will be likely to be bought by the same customer. Hence, we generate the input network from the full network by randomly removing 10\% of product-transaction edges; we use the full network as the target network for this task.

For the O-A task, we are to simulate the same practical case as the ICFCA dataset. We use the same input network and target network as those used in the O-O task of this dataset.

\textbf{Keyword-Paper}: The Keyword-Paper dataset is an original dataset generated by us -- the objects represent the keywords, the attributes represent the publications, and each edge represents the paper has the keyword. It is generated from the DBLP dump on 31st Jan 2023. From the dump, we select the top 162 most frequent keywords and all 5640 publications after 1st Jan 2010 to create the keyword-paper network. 

For the O-O task, we are to simulate the practical case in which we wish to predict potentially related keywords, which may give inspiration for new research. For example, while ``BERT'' and ``FCA'' were never used as keywords of the same paper before, if they are predicted to be related, researchers may get inspired and draft a new study similar to ours. The generation of the input network and target network are similar to that of the ICFCA dataset, and the date for the history network is also set to 31st Dec 2015.

For the O-A task, we are to simulate the same practical case as the previous two datasets. We also use the same way to generate the input and target networks as the way we used in the O-A tasks of the previous two datasets. Note that in this dataset, after removing 10\% of edges, some attributes will have no edge connecting to them, so the number of attributes in the networks used for the O-A task is smaller than that used for the O-O task.

\subsection{Evaluation}

To give a fair and comprehensive evaluation, we use the following three measures: \textit{$F_1$ score}, \textit{AUC score}, and \textit{AUPR score}. All three scores are estimated with the four basic values: TP, TN, FP, and FN. TP represents the number of samples that are positive and are predicted positive. FN represents the number of samples that are negative but are falsely predicted to be positive. TN represents the number of samples that are negative but are predicted negative. FN represents the number of samples that are positive but falsely predicted negative.

The $F_1$ score is the harmonic mean of the \textit{precision} and \textit{recall}. Precision, recall, and $F_1$ are estimated as follows: 
\begin{equation}
\begin{aligned}
    Percision &:= \frac{TP}{TP + FP}, \\
    Recall &:= \frac{TP}{TP + FN}, \\
    F_1 &:= \frac{2}{recall^{-1} + percision^{-1}}.
\end{aligned}
\end{equation}

The $F_1$ value may vary as we change the threshold for the prediction score. Hence, in this research, for each test case, we try 20 different thresholds with the following procedure -- the initial threshold is set to 0, and at each trial, we add the threshold by 0.05. After all trials, we report the highest $F_1$ value we get.

The AUC (Area Under the Curve) score is estimated by computing the area under the ROC (Receiver Operating Characteristic) curve. The ROC curve is created by plotting the true positive rate (TPR) against the false positive rate (FPR) at various threshold settings. TPR and FPR are estimated as follows:
\begin{equation}
\begin{aligned}
    TPR &:= \frac{TP}{TP + FN}, \\
    FPR &:= \frac{FP}{TN + FP}.
\end{aligned}
\end{equation}
The AUPR (Area Under the Precision-Recall Curve) score is estimated by computing the area under the Precision-Recall curve, which is created by plotting the precision rate against the \textit{recall} rate at various threshold settings.

\subsection{The Experiment of the O-O Task}

We evaluate the performance of BERT4FCA in the O-O task on the three datasets. We compare BET4FCA with two FCA-based methods, object2vec-CBoW and object2vec-SG, and a classic non-FCA-based method, node2vec, which has proven the capability to predict links in huge networks~\cite{node2vec}. The results are shown in Table~\ref{OO}.

\begin{table}[!htbp]
    \centering
    \begin{tabular}{c|ccc|ccc|ccc}
    \hline
         \multirow{2}{*}{Method} &  \multicolumn{3}{c|}{ICFCA} &  \multicolumn{3}{c|}{BMS-POS} & \multicolumn{3}{c}{Keyword-Paper} \\
    \cline{2-10}
                                 & $F_1$ & AUC & AUPR &  $F_1$ & AUC & AUPR & $F_1$ & AUC & AUPR \\
    \hline
         O2V-CBoW         & 0.686 & 0.691 & 0.672 & 0.690 & 0.637 & 0.560 & 0.463 & 0.504 & 0.518     \\
         O2V-SG           & 0.652 & 0.497 & 0.519 & 0.676 & 0.649 & 0.577 & 0.236 & 0.290 & 0.388    \\
         Node2Vec                & 0.587 & 0.703 & 0.751 & 0.870 & 0.946 & 0.945 & 0.438 & 0.647 & 0.635 \\
         BERT4FCA                & \textbf{0.781} & \textbf{0.877} & \textbf{0.896} & \textbf{0.871} & \textbf{0.964} & \textbf{0.963} & \textbf{0.800} & \textbf{0.911} & \textbf{0.908} \\
    \hline
    \end{tabular}
    \caption{The results for the O-O tasks. O2V stands for object2vec.}
    \label{OO}
\end{table}
The results show that BERT4FCA outperforms the other three models across all three datasets. We have also discovered that although node2vec has achieved high scores close to our method on the BMS-POS dataset, its performances on the other two datasets are much lower than our method. This shows the stability of our method, \textit{i.e.}, we can have a stable and high performance across different datasets.

\subsection{The Experiment of the O-A Task}

We evaluate the performance of BERT4FCA in the O-A task on the three datasets. We compare BET4FCA with an FCA-based method, Structure Hole~\cite{missbin}, and two widely-used non-FCA-based methods -- node2Vec and matrix factorization with singular value decomposition (MF-SVD). The result of object-attribute link prediction is reported in Table~\ref{OA}. 

\begin{table}[!htbp]
    \centering
    \begin{tabular}{c|ccc|ccc|ccc}
    \hline
         \multirow{2}{*}{Method} &  \multicolumn{3}{c|}{ICFCA} &  \multicolumn{3}{c|}{BMS-POS} & \multicolumn{3}{c}{Keyword-Paper} \\
    \cline{2-10}
                                 & $F_1$ & AUC & AUPR &  $F_1$ & AUC & AUPR & $F_1$ & AUC & AUPR \\
    \hline
         Structure Hole        & 0.018 & 0.000 & 0.000 & 0.226 & 0.000 & 0.000 & 0.005 & 0.000 & 0.000      \\
         Node2vec              & 0.671 & 0.800 & 0.764 & 0.612 & 0.793 & 0.755 & 0.661 & 0.747 & 0.697   \\
         MF-SVD              & \textbf{0.749} & 0.798 & 0.508 & \textbf{0.824} & \textbf{0.892} & 0.641 & 0.733 & 0.693 & 0.424 \\
         BERT4FCA              & 0.741 & \textbf{0.812} & \textbf{0.777} & 0.725 & 0.823 & \textbf{0.788} & \textbf{0.744} & \textbf{0.765} & \textbf{0.711} \\
    \hline
    \end{tabular}
    \caption{The results for the O-A tasks.}
    \label{OA}
\end{table}

From the results, we can see that BERT4FCA outperforms Structural Hole and node2vec across all three datasets. For MF-SVD, although it has achieved higher scores than BERT4FCA in some metrics, its overall performance is still considered lower than BERT4FCA for the following reasons. In the ICFCA dataset, the $F_1$ scores of BERT4FCA and MF are close, while the AUC score and especially the AUPR score of BERT4FCA are better than MF-SVD; in the BMS-POS dataset, MF-SVD has achieved higher $F_1$ and AUC scores, but its AUPR score is much lower than BERT4FCA. According to previous research, the AUPR score gives more weight to the positive samples~\cite{aucaupr}. A low AUPR score indicates that the model's high prediction scores do not correlate well with being positive class, which suggests that the model has difficulty achieving high precision~\cite{aucaupr}. Since in both tasks of bipartite link prediction, we focus on predicting the positive samples but not the negative samples~\cite{bipar1,matrixnega}, so if a model has such a problem, it is considered to have a low performance. 

We have also found that the performance of the other FCA-based method, Structure Hole, is notably lower on these three datasets than the datasets used in~\cite{missbin}. As analyzed before, Structure Hole is a rule-based method that uses the same simple rule to extract information from the concept lattices and make predictions on all datasets, which will be highly likely to have lower performance on some datasets because not all datasets fit the rule well, and our experimental results just proved it. This again shows the importance of developing a method like our BERT4FCA that can automatically capture the information from concept lattices using statistical machine-learning techniques.

\subsection{Supplementary Experiments}
To verify if our original mechanisms are functioning well, we conduct two supplementary experiments -- the first one checks if our model can indeed learn the neighboring relations of formal concepts; the second one checks if the information in concept lattices our model learned is indeed helpful for making link predictions.

In the first supplementary experiment, we only use 80\% of the training samples to pre-train both the object model and the attribute model on three datasets. The remaining 20\% of the original training samples are kept as test samples for the NCP task -- these intents/extents pairs are not presented in the training samples, so if our model can correctly predict whether they are neighbors or not, it should indicate that our method has learned the structure of concept lattice well. The results are shown in Table~\ref{NCP}.

\begin{table}[!htbp]
    \centering
    \begin{tabular}{c|c|ccc}
    \hline
         Dataset & Object/Attribute & $F_1$ & AUC & AUPR \\
    \hline
        \multirow{2}{*}{ICFCA} & Object & 0.903 & 0.896 & 0.842 \\
    \cline{2-2}
                               & Attribute & 0.868 & 0.842 & 0.785 \\
    \hline
        \multirow{2}{*}{BMS-POS} & Object & 0.993 & 0.993 & 0.987 \\
    \cline{2-2}
                                 & Attribute & 0.978 & 0.978 & 0.963 \\
    \hline
        \multirow{2}{*}{Keyword-Paper} & Object & 0.965 & 0.966 & 0.934 \\
    \cline{2-2}
                                       & Attribute & 0.900 & 0.902 & 0.850 \\
    \hline
    \end{tabular}
    \caption{The results for the first supplementary experiment.}
    \label{NCP}
\end{table}

The results suggest that BERT4FCA indeed learned the neighboring relations from the concept lattice on all three datasets well. We have also noticed that the results of the object models are generally better than the attribute models across all datasets. We analyze it because, in these three datasets, the average lengths of intents are longer than that of the extents, making it potentially more challenging to effectively capture the neighboring relations of intents. 

In the second supplementary experiment, we check if the information we learned from concept lattices indeed contributes to better link prediction results. Since the information of the concept lattices is learned in the NCP and MTM tasks in the pre-training step, in this experiment, we skip the pre-training step and directly fine-tune the models with randomly initialized weights. By comparing the results with that of our full method, we will know whether our specially designed tasks for learning the information of concept lattices are functioning well. The results are shown in Tables~\ref{OO-concept-infp} and~\ref{OA-concept-infp}.

\begin{table}[!htbp]
    \centering
    \begin{tabular}{c|ccc|ccc|ccc}
    \hline
         \multirow{2}{*}{Method}    &  \multicolumn{3}{c|}{ICFCA} &  \multicolumn{3}{c|}{BMS-POS} & \multicolumn{3}{c}{Keyword-Paper} \\
    \cline{2-10}
                                    & $F_1$ & AUC & AUPR &  $F_1$ & AUC & AUPR & $F_1$ & AUC & AUPR \\
    \hline
         BERT4FCA-NC & 0.718 & 0.810 & 0.815 & 0.865 & 0.961 & 0.958 & 0.735 & 0.866 & 0.858 \\
         BERT4FCA                   & 0.781 & 0.877 & 0.896 & 0.871 & 0.964 & 0.963 & 0.800 & 0.911 & 0.908 \\
    \hline
    \end{tabular}
    \caption{The results for the O-O tasks of the second supplementary experiment.}
    \label{OO-concept-infp}
\end{table}

\begin{table}[!htbp]
    \centering
    \begin{tabular}{c|ccc|ccc|ccc}
    \hline
         \multirow{2}{*}{Method}    &  \multicolumn{3}{c|}{ICFCA} &  \multicolumn{3}{c|}{BMS-POS} & \multicolumn{3}{c}{Keyword-Paper} \\
    \cline{2-10}
                                    & $F_1$ & AUC & AUPR &  $F_1$ & AUC & AUPR & $F_1$ & AUC & AUPR \\
    \hline
         BERT4FCA-NC & 0.723 & 0.801 & 0.766 & 0.701 & 0.800 & 0.760 & 0.724 & 0.748 & 0.707 \\
         BERT4FCA    & 0.741 & 0.812 & 0.777 & 0.725 & 0.823 & 0.788 & 0.744 & 0.765 & 0.711 \\
    \hline
    \end{tabular}
    \caption{The results for the O-A tasks of the second supplementary experiment.}
    \label{OA-concept-infp}
\end{table}

The results show that learning more information from concept lattices indeed contributes to the improvement of performances in both tasks of bipartite link prediction, while the degree of improvement varies across different datasets and tasks. The overall degree of improvement in O-O tasks is higher than that in the O-A task. We analyze it because compared to the attribute models, the object models better capture the information on concept lattices, such as the neighboring relations, as demonstrated in the first supplementary experiment. Also, we find that in the O-O task on the BMS-POS dataset, learning the information from concept lattices contributes to the lowest degree of improvement. This may be because the performance of our model without learning the information is already sufficiently high, as the AUC and AUPR scores have reached 96\%. In such a case, it is considered hard to improve the performance further, even with an effective mechanism.

\section{Conclusion and Future Work}

In this paper, we proposed BERT4FCA, a novel FCA-based method for bipartite link prediction. It can learn more information from concept lattices and then use it to make bipartite link prediction. The experimental results demonstrated that our methods outperform previous FCA-based methods like object2Vec and non-FCA-based classic methods like MF and node2vec in both the O-O and the O-A tasks on three different datasets. The results have shown that our method is stabler than previous methods. We also demonstrated by extra experiments that neighboring relations between maximal bi-cliques are well learned by the model as expected and that such information contributes to better link prediction results. Furthermore, we have shown that BERT4FCA provides a general framework for employing BERT to learn the information extracted by FCA. Hence, in addition to bipartite link prediction, we believe BERT4FCA can also be further applied to a broader range of real-world tasks. We plan to explore the potential applications in the future.

However, we have also found some points of our method that may be further refined and improved. First, limited by computational resources, we were unable to pre-train an all-in-one model for both objects and attributes in this research. If such a model can be trained well, it will be more convenient as all downstream tasks will only need one single pre-trained model. Second, although BERT4FCA outperforms previous methods in prediction results, the pre-training step is time-consuming, especially when dealing with datasets with a large number of formal concepts. Although there have been methods for reducing the number of formal concepts in concept lattices, it is uncertain whether training on a reduced concept lattice will result in lower performances. If so, we may need to work out a technique to make a trade-off between the training time and the performance. We plan to address these two points in our future work.

\section*{Declarations}

\textbf{Conflict of Interest} There is no conflict of interest.

\bibliography{sn-bibliography}% common bib file

\end{document}